\documentclass[letterpaper, 10 pt, conference]{ieeeconf}  

\IEEEoverridecommandlockouts                              

\title{\LARGE \bf
Infinite-Dimensional Adaptive Boundary Observer for Inner-Domain Temperature Estimation of 3D Electrosurgical Processes using Surface Thermography Sensing
}

\author{Hamza~El-Kebir$^{1}$, Junren~Ran$^{2}$, Martin Ostoja-Starzewski$^{3}$, Richard Berlin$^{4}$, \\ Joseph~Bentsman$^{2}$, and Leonardo~P.~Chamorro$^{2}$
\thanks{$^{1}$H. El-Kebir is with the Department
of Aerospace Engineering, University of Illinois Urbana-Champaign, Urbana,
IL, 61801 USA
{\tt\small elkebir2@illinois.edu}}%
\thanks{$^{2}$J. Ran, J. Bentsman, and L. P. Chamorro are with the Department
of Mechanical Science and Engineering, University of Illinois Urbana-Champaign, Urbana,
IL, 61801 USA
{\tt\small \{jran2,jbentsma\}@illinois.edu}}%
\thanks{$^{3}$M. Ostoja-Starzewski is with the Department
of Mechanical Science and Engineering, the Beckman Institute, and the Institute for Condensed Matter Theory, University of Illinois Urbana-Champaign, Urbana, IL, 61801 USA
        {\tt\small martinos@illinois.edu}}%
\thanks{$^{4}$R. Berlin is with the Department of Trauma Surgery, Carle Hospital and the University of Illinois Urbana-Champaign, Urbana, IL 61801 USA.}%
\thanks{Research reported in this publication was supported by the National Institute of Biomedical Imaging and Bioengineering of the National Institutes of Health under award number R01EB029766. The content is solely the responsibility of the authors and does not necessarily represent the official views of the National Institutes of Health.}
}

\let\labelindent\relax

\usepackage{amsmath}
\usepackage{mathrsfs}
\usepackage{graphicx}
\usepackage[inline]{enumitem}

\usepackage{bm}

\usepackage{tikz}
    \usetikzlibrary{perspective}
    \usetikzlibrary{3d}
    \usetikzlibrary{arrows.meta}

\usepackage{amsthm}

\newtheorem{theorem}{Theorem}

\newtheorem{lemma}{Lemma}
\newtheorem{conjecture}{Conjecture}
\theoremstyle{remark}
\newtheorem{remark}{Remark}

\usepackage[T1]{fontenc}
\usepackage{calligra}

\usepackage{stix}

\usepackage{dsfont}
\usepackage{hyperref}
\usepackage[inline]{enumitem}
\usepackage{graphicx}
\usepackage{caption}
\usepackage{subcaption }
\usepackage{tabularx}
\usepackage{booktabs}

\usepackage{scalerel}

\begin{document}

\maketitle
\thispagestyle{empty}
\pagestyle{empty}

\begin{abstract}
We present a novel 3D adaptive observer framework for use in the determination of subsurface organic tissue temperatures in electrosurgery. The observer structure leverages pointwise 2D surface temperature readings obtained from a real-time infrared thermographer for both parameter estimation and temperature field observation. We introduce a novel approach to decoupled parameter adaptation and estimation, wherein the parameter estimation can run in real-time, while the observer loop runs on a slower time scale. To achieve this, we introduce a novel parameter estimation method known as attention-based noise-robust averaging, in which surface thermography time series are used to directly estimate the tissue's diffusivity. Our observer contains a real-time parameter adaptation component based on this diffusivity adaptation law, as well as a Luenberger-type corrector based on the sensed surface temperature. In this work, we also present a novel model structure adapted to the setting of robotic surgery, wherein we model the electrosurgical heat distribution as a compactly supported magnitude- and velocity-controlled heat source involving a new nonlinear input mapping. We demonstrate satisfactory performance of the adaptive observer in simulation, using real-life experimental ex vivo porcine tissue data.
\end{abstract}


%
\IEEEpeerreviewmaketitle

\section{Introduction}
%
%
%
%

Over the past decades, the advantages of robot-assisted surgery over conventional laparoscopy have become increasingly apparent \cite{Lanfranco2004}. As the robotic surgery performance demands grow, so does the need for a deeper understanding of the physical phenomena governing both the tools and the tissue treated, as well as the control laws for the precise attainment of the surgical objectives.

\begin{figure}[t]
    \includegraphics[width=\linewidth]{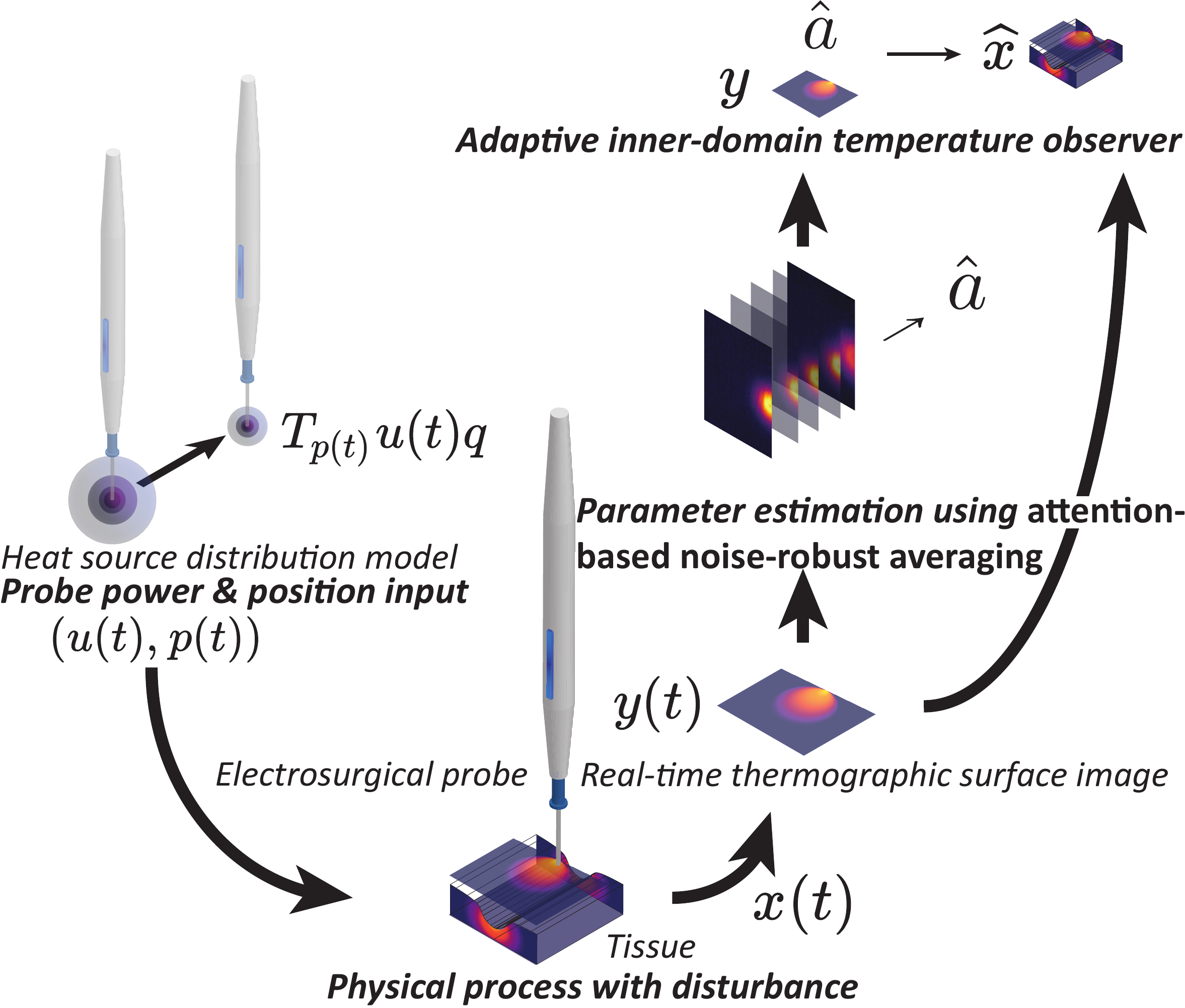}
    \caption{Illustration of the electrosurgical process as it relates to the adaptive inner-domain temperature observer presented in this work. Given the electrosurgical probe power and position commands, as well as a surface thermography image, we estimate the thermal diffusivity and employ an adaptive observer law to obtain the inner-domain temperature field.}
    \label{fig:electrosurgery}
\end{figure}

Electrosurgery relies on the use of high power density radio frequency currents to actively heat organic tissue, allowing it to be denatured, coagulated, desiccated, fulgurated, or incised \cite{Palanker2008}. One of the key advantages of electrosurgery is its capability of simultaneous cutting and coagulation, providing blood stoppage for complex surgical tasks. Since this technique allows for precise ablation of tissue with very little collateral damage ($\sim$100--400 $\mu$m), it is commonly used in practice, with over half of the surgical procedures employing it \cite{Palanker2008}. An illustration of the electrosurgical process along with the observer setup considered in this work is shown in Fig.~\ref{fig:electrosurgery}.

Electrosurgical control can be posed as a boundary control problem characterized by a controllable heat flux, as dictated by the power setting, and cathode position actuation. In previous work \cite{El-Kebir2021c, El-Kebir2021d}, the authors have considered a 1D setting in which the tissue thermodynamics are modeled using a Stefan problem formulation, where a homogeneous substance undergoing a moving phase change from \textit{virgin} to a \textit{denatured} tissue is considered. In \cite{El-Kebir2021c} a stationary probe was considered, with a novel approach to non-collocated thermal sensing being introduced. The model in \cite{El-Kebir2021c} was applied as part of a real-life planar cutting task using a hypermodel-based model predictive control algorithm in \cite{El-Kebir2021d}. However, to guarantee a tight attainment of the control objective in the case of a moving probe location, which is essential in practice \cite{Palanker2008}, considering the process in 3D becomes a necessity. In this work, we consider a spatially distributed controlled moving heat source, while accounting for chemical reactions by incorporating a bounded disturbance. In this setting, as the first approximation, a homogeneous Fourier heat equation is employed, with a moving heat volumetric source that has compact spatial support. In this work, we develop a 3D observer structure based on surface temperatures with the goal of estimating the subsurface temperature distribution; the latter goal is essential in assessing and preventing subsurface tissue damage. In practice, these inner-domains temperatures cannot be sensed in a noninvasive fashion, thereby significantly limiting any model-based control efforts for safe autonomous electrosurgery. In addition to proving convergence under bounded heat input disturbances in the case of known thermophysical parameters, we also introduce a novel real-time parameter adaptation loop in which \emph{only} surface temperature measurements are used. This provides the first adaptive surface-temperature-based observer for electrosurgical processes.

In the context of biomechanics and electrosurgery, the first works dealing with temperature distributions in nonhomogeneous tissue models considered the thermal response of layered media when subjected to microwave excitation using microwave radiometry \cite{Bardati1983}, building on earlier work that considered homogeneous tissue models \cite{Erez1980}. In all of these works, experimental verification rested on the use of surface temperatures measurements; indeed, obtaining inner-domain temperatures is not possible without invasively altering the heat transfer process, e.g., by inserting thermocouples. Recent work \cite{Ledwon2022} has used multi-angle surface thermography images to estimate the tomographic, or volumetric, temperature field. In \cite{Ledwon2022}, an application-specific deep neural network model was trained to deduce the inner-domain temperatures from the surface temperatures after being trained on synthetic data based on an unforced Fourier heat transfer model with constant thermal diffusivity throughout the entire domain. Given this data, such a model is only capable of estimating inner-domain temperatures for experiments that involve a free thermal response and have the same thermophysical parameters, boundary conditions, and geometry. In the context of electrosurgery, it is essential that an external heat input be accounted for, as well as to have an ability to obtain subsurface temperatures for different tissue types and geometries. It is then natural to consider the infinite-dimensional observers based on the surface temperature feedback, as is done in this work. 

In constructing model-based observers, it is essential to have a sufficiently accurate model of the thermodynamics. While the thermal properties of the tissue can readily be obtained in a laboratory environment using differential scanning calorimetry (DSC) and  thermogravimetric analysis (TGA) \cite{El-Kebir2021c}, these methods are costly and time-consuming, and only provide localized estimates of the tissue properties. More importantly, such methods are necessarily invasive, since they require biopsic samples to be extracted which can often only be used for one form of analysis at a time. Circumventing these issues, which extend to nondestructive testing of materials as well, in the context of subsurface material characterization a technique known as dynamic thermal tomography (DTT) has been introduced by Vavilov \emph{et al.} \cite{Vavilov1992}, and has recently seen use in medical applications \cite{Toivanen2014, Toivanen2017}. In DTT, thermographic images are captured at different areas on the surface, with the goal of estimating local thermophysical characteristics (thermal conductivity, heat capacity, heat transfer coefficients, etc.) by exciting dynamic thermal responses using embedded collocated heating elements. Using the obtained volumetrically varying parameter estimates, it is possible to formulate a model PDE and obtain the subsurface temperatures as was done in \cite{Toivanen2014}. An early approach to this method for use in biomedical applications is presented in \cite{Chen1985}, but it relies on Fourier series expansions. A major drawback of DTT is the fact that it is an invasive tomography method, unlike other tomographic imaging techniques such as magnetic resonance imaging. In this work, we introduce a non-invasive real-time parameter adaptation framework based solely on the surface temperature. To enable the latter we propose a novel thermography-based parameter estimation method which we call \emph{attention-based noise-robust averaging} (ANRA).

The paper has the following structure. The boundary control problem of interest along with the related mathematical model are presented in Sec.~\ref{sec:Problem Formulation}. Sec.~\ref{sec:Observer Design} presents a non-collocated surface-temperature-based observer structure for electrosurgical action that is asymptotically convergent in the case of known thermophysical parameters. Sec.~\ref{sec:Adaptive Observer Design} presents a novel adaptive observer structure, in which the thermophysical parameters are directly estimated from the real-time surface thermography. Sec.~\ref{sec:Numerical Simulation} compares the performance of both observer structures through numerical simulation. Conclusions are drawn in Sec.~\ref{sec:Conclusion}. Formal proofs will be presented in a separate publication.

\section{Problem Formulation}\label{sec:Problem Formulation}

In this work, we consider the following thermodynamic model on an $n$-dimensional compact domain $\Omega \subseteq \mathbb{R}^n$, based on a standard model from the literature \cite[Eq.~1]{Singh2020}:
\begin{equation}\label{eq:nominal system}
    \dot{z}(t) = \begin{bmatrix}
        \dot{x}(t) \\
        \dot{p}(t)
    \end{bmatrix} =
    \begin{bmatrix}
        a \nabla^2 x(t) + T_{p(t)} u(t) q + w(t) \\
        v(t)
    \end{bmatrix},
\end{equation}
where $a \in \mathbb{R}_+$ denotes the thermal diffusivity, and $\nabla^2$ is the Laplacian operator. $x(t)$ denotes the temperature field at time $t$, such that $x(t) \in H^1 (\Omega)$; by $x(t, \eta)$ we denote $x(t)(\eta)$ for $\eta \in \Omega$. The heat source position is denoted by $p(t) \in \Omega$, which is used as a parameter in the functional $T_p$ defined as
\begin{equation}
\begin{split}
    T_{p} : H^1 (\Omega) &\to H^1 (\Omega), \quad \forall p \in \Omega, \\
    T_{p} q(\eta) &\mapsto \chi_{\Omega}(\eta) q(\eta - p), \quad \forall \eta \in \Omega, q \in H^1 (\Omega),
\end{split}
\end{equation}
where $\chi_\Omega$ is the indicator function on the set $\Omega$. In our model, $q \in H^1(\Omega)$ denotes a normalized heat source term, which is scaled by the control input $u(t) \in \mathbb{R}_+$. The heat input is perturbed by $w(t) \in \mathcal{W} := \{ w \in H^1 (\Omega) : \Vert w \Vert_{L^2(\Omega)} \leq \epsilon_{w, 2}, w \in C^{0, \gamma}(\Omega) \}$, which is bounded in the $L^2 (\Omega)$-norm by known $\epsilon_{w, 2} \geq 0$, and is $\gamma$-H\"older continuous for $\gamma = 1/2$. Finally, $v \in C^0 (\mathbb{R}^n)$ is the known controlled heat source velocity, which is such that $p(t) := p(0) + \int_0^t v(\tau) \ \mathrm{d}\tau \in \Omega$ for all $t \in [0, \infty)$. We denote the full state by $z \in H^1 (\Omega) \times C^1(\Omega)$.

In this work, we assume having access to thermographer data captured on one of the boundaries of $\Omega$, which we denote by $\Gamma := \{\eta \in \Omega : \eta_n = 0\}$. In a three-dimensional setting, this would be the top surface of a slab of material. Furthermore, we consider that temperatures are sampled on a finite Cartesian grid with uniform spacing $\Delta \eta > 0$, denoted by $\hat{\Gamma}$. In this work, we consider $\mathrm{dim} \ \hat{\Gamma} = M < \infty$, such that $\hat{\Gamma} := \{\eta^{(i)}\}_{i=1}^M$.

Let $C : H^1 (\Omega) \to \mathbb{R}^M$ be the output map, defined as $C x(t) := \left(x(t, \eta^{(i)})\right)_{i=1}^M$. The output is denoted by $y(t) := C x(t)$. We introduce a map $C^*$ from $\mathbb{R}^M$ to $H^1 (\Omega)$, which will be used as part of a Luenberger observer structure in the following section:
\begin{equation*}
    \left( C^* y(t) \right)(\eta) := \sum_{i=1}^M y_i \delta(\eta - \eta^{(i)}) = \sum_{i=1}^M x(t, \eta^{(i)}) \delta(\eta - \eta^{(i)}),
\end{equation*}
where $\delta$ is the Dirac delta. In the remainder of this work, we make the following assumptions:

\begin{enumerate}[label=(A\arabic*)]
    \item The initial condition $x(0) \in H^1 (\Omega)$ has known constant temperature, i.e., $x(0, \eta) = x_0 \chi_{\Omega}(\eta)$ for all $\eta \in \Omega$ for some known $x_0 \in \mathbb{R}$.
    \item $\inf u(t) \geq 0$ and $\sup u(t) < \infty$.
    \item On the boundary of $\Omega$, $\partial \Omega$, a trivial Neumann boundary condition holds, i.e., $\nabla x(t) |_{\partial\Omega} \equiv 0$.
    \item The disturbance $w(t)$ is bounded in the $L^2$-norm $\Vert w(t) \Vert_{L^2(\Omega)} \leq \epsilon_{w, 2}$ for all $t \in [0, \infty)$, for known $\epsilon_{w, 2} \geq 0$. In addition, $w(t)$ is $\gamma$-H\"older for $\gamma = 1/2$, with the H\"older coefficient being upper-bounded by a known constant $C_w \geq 0$. Finally, $w \in C^0 (\mathcal{W})$, i.e., $w$ is continuous in time.
    \item Heat source velocity $v \in C^0 (\mathbb{R}^n)$ is continuous and bounded, such that $p(t) := p(0) + \int_0^t v(\tau) \ \mathrm{d}\tau \in \Omega$ for all $t \in [0, \infty)$.
\end{enumerate}

As can be seen in \eqref{eq:nominal system}, we assume that there are no errors in the probe position. In the remainder of this work, we consider control input pair $(u(t), v(t))$ to be given for all $t \in [0, \infty)$; we also assume that there is no error in the control inputs.

\begin{remark}
It will be shown in Sec.~\ref{sec:Numerical Simulation} that Assumption~(A1) does not necessarily need to hold. Assumption~(A2) requires boundedness and non-negativity of the control signal, which follows from physical limitations in electrosurgery (i.e., heat can only be supplied by the electrosurgical pencil, and only a finite power can be supplied). Assumption~(A3) is standard and can be argued to hold for sufficiently large domains, where the far-field temperature is constant (i.e., in thermodynamic equilibrium).

Assumption~(A4) puts constraints on the heat source disturbance. First, it should be noted that the $L^2$-norm bound imposed by $\epsilon_{w,2} \geq 0$ requires that the disturbance signal is of bounded energy \textit{in space}, but not in time. This implies that the disturbance could be persistent across the entire system run, with no ultimate boundedness being imposed. Furthermore, the $1/2$-H\"older coefficient bound $C_w \geq 0$ imposes a constraint on the sharpness of the disturbance; in particular, this constraint is more general than a Lipschitz bound. While white noise cannot be captured as part of this assumption, wide-band non-decaying signals can be. This $L^2$-norm bound can be found by considering unmodeled dynamics such as tissue perfusion, and accounting for the magnitude of these terms based on \cite{Singh2020}.
\end{remark}

%

\section{Observer Design}\label{sec:Observer Design}

We consider the following observer system:
\begin{equation}\label{eq:observer}
\dot{\hat{x}}(t) = a_0 \nabla^2 \hat{x}(t) + T_{p(t)} u(t) q + C^* G(t) (y(t) - \hat{y}(t)),
\end{equation}
where $G(t) \in \mathbb{R}^{M \times M}$ is the time-varying observer gain, which is diagonal and positive definite for all $t \geq 0$, such that at least one of the $g_i(t) := G_{ii} (t)$ satisfies
\begin{equation}\label{eq:gain bound}
\begin{split}
    g_i(t) \geq \ &\epsilon_{w, 2} \ m(\Omega)^{1/2} \ \tilde{x}_{s,i}^{-2}(t) \\
    &\times \left( \max_{\eta \in \Omega} \min_{i = 1,\ldots,M} \left[ |\tilde{x}_{s,i}(t)| + C_\gamma(t) \Vert \eta^{(i)} - \eta \Vert^\gamma \right] \right),
\end{split}
\end{equation}
    where $m(\Omega)$ is the volume of the domain $\Omega$, and
\begin{equation}\label{eq:holder bound}
    C_\gamma (t) = \frac{C_w}{n a} (\exp(n a \ t) - 1).
\end{equation}

Note that $a_0$ is assumed to be known; the case of unknown thermal diffusivity, is treated in the next section. Here, the disturbance H\"older coefficient bound $C_w$ and the $L^2$-bound $\epsilon_{w, 2}$ are known based on Assumption~(A4). Note that, as mentioned in the previous section, we assume that $p \in C^1 (\Omega)$ is known without error. We start by considering asymptotic error convergence for the case where $a_0 = a$.

\subsection*{Observer Error Convergence}

\begin{theorem}\label{thm:Estimation Error Convergence}
    Consider the observer given by \eqref{eq:observer}--\eqref{eq:holder bound}. Provided that assumptions (A1)--(A4) hold, the temperature estimation error converges to zero.
\end{theorem}

\begin{remark}
The lower bound on the gain given by \eqref{eq:gain bound} and \eqref{eq:holder bound} can be computed directly given the internal observer state $\hat{x}(t)$, the observed surface temperature $y(t)$, and the known bounds $\epsilon_{w, 2}$ and $C_w$ from (A4). Given the assumptions, $\tilde{x}_{s,i}(t)$ is continuous, which in turn implies that the lower bound on the gain, \eqref{eq:gain bound}, is continuous.

Given the evolution of $C_\gamma(t)$ from \eqref{eq:holder bound}, it can be seen that the lower gain bound is increasing in time, but is bounded for finite $t$. In practice, a reasonably large initial choice of gain will result in \eqref{eq:gain bound} being satisfied during some time interval $[0, T]$ for $T \in \mathbb{R}_+$.

Theoretically, the elements $g_i(t)$ could be taken arbitrarily large, since no sensor noise is considered as part of this work; this would result in an instantaneous correction of the temperature error at each sensing location. In practice, one would not want to increase the gain too much, since this results in amplification of unwanted sensing noise.
\end{remark}

\section{Adaptive Observer Design}\label{sec:Adaptive Observer Design}

We now consider the case where $a_0 \neq a$, with unknown $a$. Given that the true diffusivity is unknown, we introduce the following adaptive observer structure:
\begin{equation}\label{eq:adaptive observer}
\begin{split}
    \dot{\hat{x}}(t) &= \bar{a}(t) \nabla^2 \hat{x}(t) + T_{p(t)} u(t) q + C^* G(t) (y(t) - \hat{y}(t)),
\end{split}
\end{equation}
where $G(t)$ obeys the conditions of Thm.~\ref{thm:Estimation Error Convergence}, and $\bar{a}(t)$ is defined as
\begin{equation}\label{eq:parameter adaptation}
    \bar{a} (t) := a_0 + L \int_0^t (\hat{a}(\tau) - \bar{a} (\tau)) \text{d}\tau,
\end{equation}
where $\hat{a}(t)$ is the instantaneous thermal diffusivity estimate obtained using the \emph{attention-based noise-robust averaging} (ANRA), introduced next. The parameter adaptation gain is denoted by $L \geq 0$. In this formulation, the parameter estimate $\bar{a}(t)$ in effect acts as an integral correction term to the Laplacian term.

\begin{figure}[htb]
\centering
\centerline{\includegraphics[width=\linewidth]{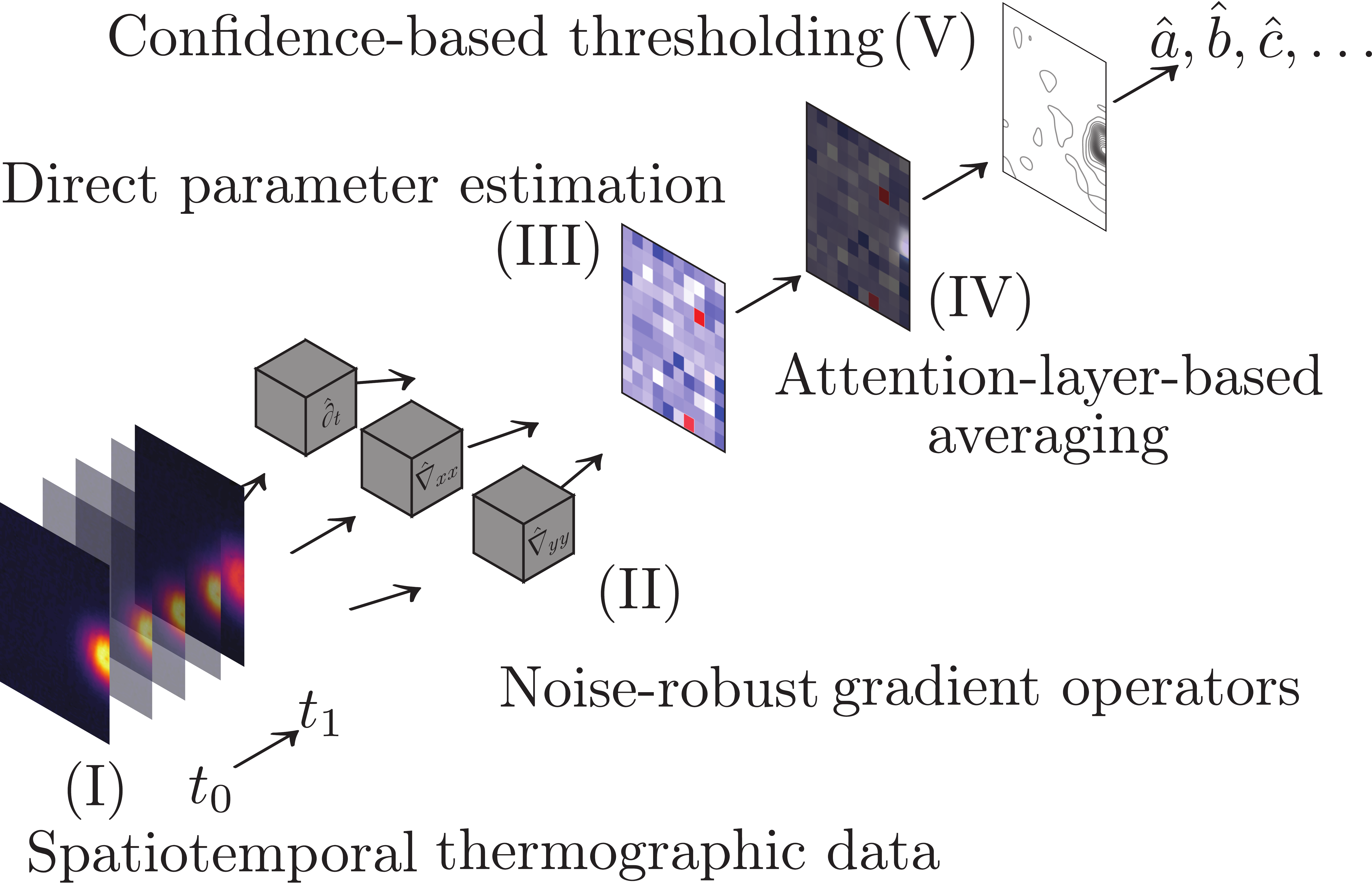}}
  \centering
  \caption{Overall \emph{attention-based noise robust averaging} (ANRA) model architecture: (I) real-time thermographic imagery is acquired; (II) gradients that appear in the PDE model are computed using noise-robust gradient operators; (III) computed gradients are used to directly estimate parameters; (IV) an attention layer based on the rate of temperature change (RTC) is used to compute weighted (spatially averaged) parameter estimates; (V) based on a confidence threshold on the RTC-field, final parameters are computed.}
  \label{fig:architecture}
\end{figure}

\begin{figure*}[t]
    \includegraphics[width=\linewidth]{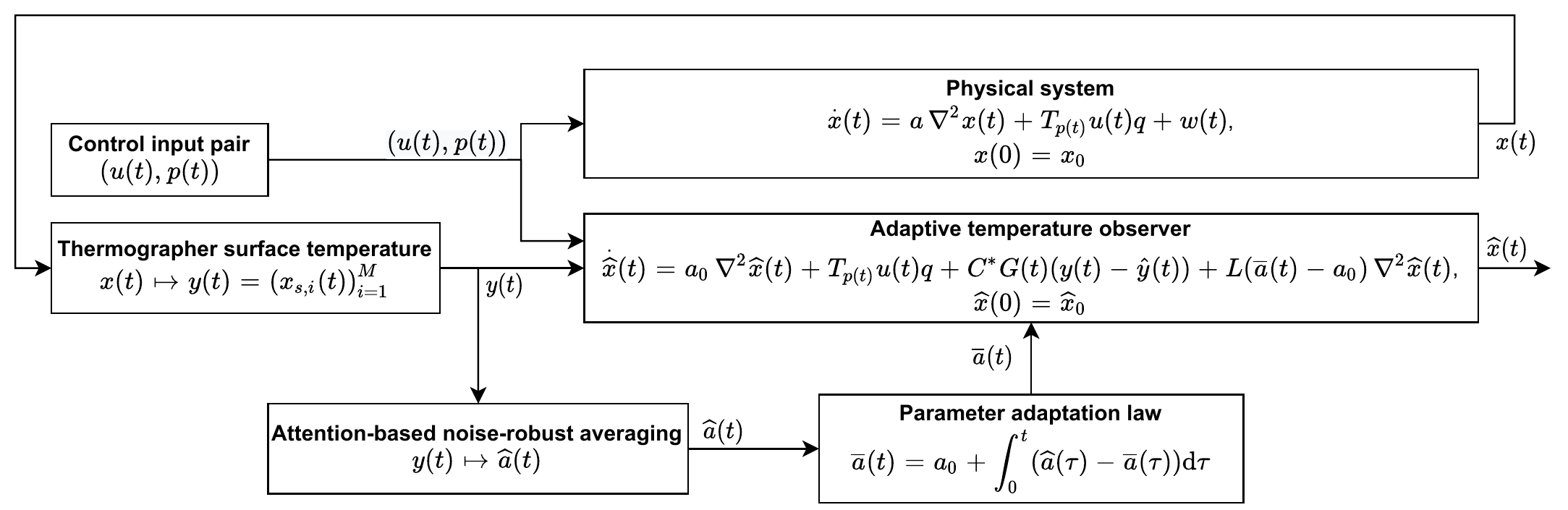}
    \caption{Block diagram of the proposed adaptive observer \eqref{eq:adaptive observer}.}
    \label{fig:adaptive observer}
\end{figure*}

\subsection*{Adaptive Noise-Robust Averaging}

To obtain $\hat{a}(t)$, we note that we may only use the surface temperature readings from the thermographer. To use the latter for our purpose we propose a novel framework further referred to as \emph{attention-based noise-robust averaging} (ANRA). ANRA relies on noise-robust gradient operators that allow for computing the Laplacian and time derivative while suppressing high-frequency noise, which would ordinarily corrupt any further data processing steps. We denote the thermal diffusivity estimate obtained by ANRA based on discrete-time surface temperature measurements by $\hat{a}[k]$, where $k$ is the current time sampling index.

Since our parameter estimation architecture depends on direct computation of the spatial and temporal temperature gradients, we wish to suppress the effects of high frequency noise, which is observed both spatially and temporally in thermographic data. For this reason, a \emph{noise-robust filter}, i.e., one that subdues high frequency noise while leaving low frequency content unattenuated, is desirable \cite{Holoborodko2008}. Pavel Holoborodko introduced noise-robust gradient operators in backward differencing form \cite{Holoborodko2008}, as well as linear filters for spatial gradients \cite{Holoborodko2009}; this family of operators is instrumental in our work. We denote noise robust finite difference operators by $\hat{\partial}$.

As part of the ANRA framework, we introduce the \emph{rate of thermal change} (RTC) field indicating the areas of the temperature field exhibiting the most activity. The RTC layer serves to indicate the points likely to be noise-driven, and allowing us to favor the high thermal activity regions. For this reason, we propose to draw the attention of the parameter estimation algorithm to areas of high RTC, using a data-driven \emph{attention layer}. The RTC field is defined as follows:
\small
\begin{equation}\label{eq:RTC definition}\begin{split}
\text{RTC}_{i,j} [k] &\leftarrow \exp(\beta |\hat{\partial}_t \bar{x}_{i,j}[k]|) - \min_{i,j} \exp(\beta |\hat{\partial}_t \bar{x}_{i,j}[k]|) \\
\text{RTC}[k] &\leftarrow \text{RTC}[k]/ \sum_{i,j} \text{RTC}_{i,j}[k],
\end{split}\end{equation}\normalsize
where $k$ denotes the sample index in time, $\bar{x}[k]$ denotes the surface temperature matrix captured by the thermographer, and $\beta \geq 1$ is a tuning parameter that increases the RTC value separation between areas of low and high rate of change as it increases; in this work, $\beta = 100$. The choice for this definition is as follows. We take the absolute value of the temperature gradient at each location and exponentiate it, so as to assign a larger positive value to a large thermal rate of change, this value scaling nonlinearly in the gradient. We then subtract the smallest element to get the minimum to be zero. Finally, we divide by the sum of all elements, to obtain a matrix whose elements sum to 1. This latter property is instrumental as shown next.

Having obtained the RTC field, we may now introduce a weighted averaging step. The instantaneous thermal diffusivity estimate is obtained as follows:
\begin{equation}\label{eq:diffusivity estimate}
    \hat{a}[k] = \sum_{i,j} \text{RTC}_{i,j}[k] \left\{ \hat{\partial}_t \bar{x}[k] / [(\hat{\partial}_{11} + \hat{\partial}_{22}) \bar{x}[k]] \right\}_{i,j}.
\end{equation}

In this step, we obtain a weighted average that is governed by the magnitude of the rate of temperature change. This is motivated by the observation that the sensor data from the areas that experience little thermal change are often dominated by noise. In light of \eqref{eq:nominal system}, thermal diffusivity estimates obtained using ANRA are only valid when the power input, $u(t)$, is zero. During surgery, such periods of inactivity frequently occur to assess the quality of an incision that was just made, thereby providing ample opportunity to compute $\hat{a}$ during these times. 
An overview of the ANRA method is shown in Fig.~\ref{fig:architecture}, and the full adaptive observer structure is summarized in Fig.~\ref{fig:adaptive observer}.

It must be noted that when obtaining the diffusivity estimates, temperature gradients in the depthward direction cannot be sensed. This is natural, since thermography cannot determine temperatures \emph{inside} a material. This does, however, lead to a discrepancy in the thermal diffusivity when compared to a correct estimate based on the full depthward temperature information. We propose a constant scaling factor on $\hat{a}[k]$ to correct for this fact, as outlined in the following lemma.

\begin{lemma}\label{lemma:diffusivity correction factor}
    Consider an infinitesimal 3D Cartesian volume of a homogeneous material, whose temperature is governed by the heat equation $\dot{\xi}(t) = a \nabla^2 \xi(t)$.
    Let the top surface of this volume be fully insulated (i.e., $\partial^2_{\hat{n}_{\text{top}}} \xi(t) = 0$). Assume that along each Cartesian axis, the second derivative of temperature along that axis is the same for each direction. Then, if the bottom surface is also modeled to be insulated, while the true system is not insulated at the bottom, the effective thermal diffusivity $a'$ will be equal to $\frac{5}{4} a$.
\end{lemma}

Using the result from Lemma~\ref{lemma:diffusivity correction factor}, it suffices to scale any thermal diffusivity estimate obtained from surface temperature measurements by $4/5$.


\begin{conjecture}\label{conj:adaptive observer convergence}
    Given the adaptation law introduced in \eqref{eq:adaptive observer}, the adaptive observer converges asymptotically, with faster convergence compared to the non-adaptive observer of Thm.~\ref{thm:Estimation Error Convergence}.
\end{conjecture}

We demonstrate Conjecture~\ref{conj:adaptive observer convergence} in the following section.

\section{Numerical Simulation}\label{sec:Numerical Simulation}

\begin{figure*}[h]
\minipage{0.32\linewidth}
    \includegraphics[trim={0cm 1.8cm 1.4cm 1.8cm},clip,width=0.9\linewidth]{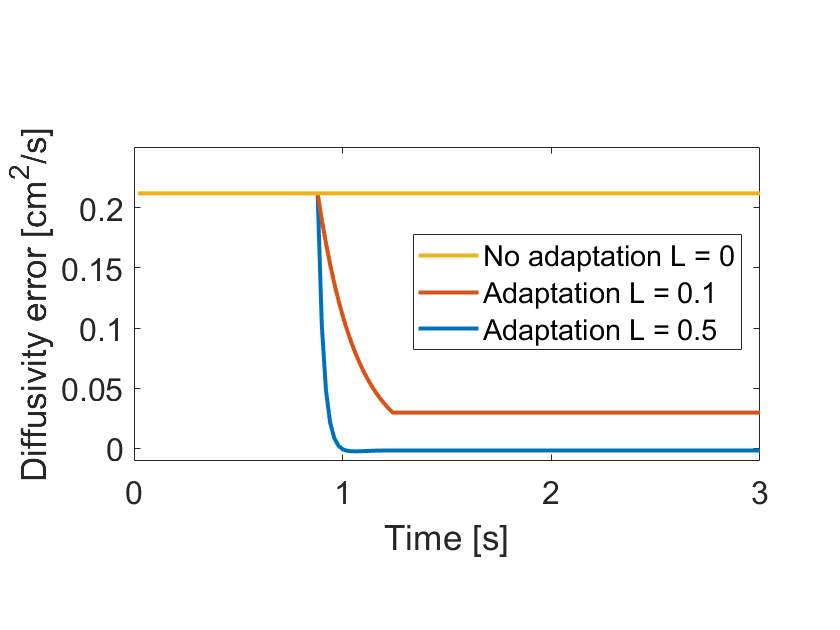}
    \caption{Parameter estimate error for different parameter adaptation gains.}        \label{fig:parameter error}
\endminipage\hfill
\minipage{0.32\linewidth}
    \includegraphics[trim={0.2cm 1.8cm 0.8cm 1.8cm},clip,width=0.9\linewidth]{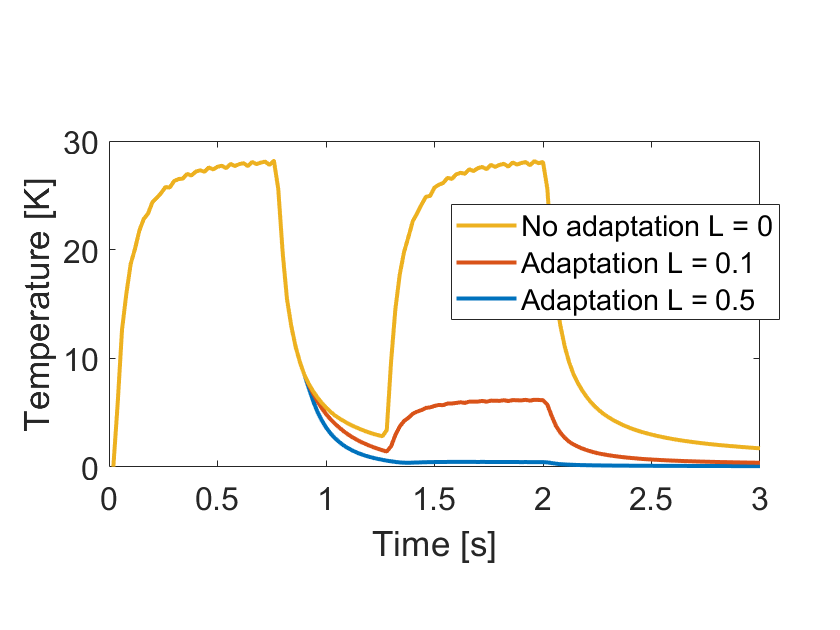}
    \caption{Maximum temperature error for different parameter adaptation gains.}
    \label{fig:error inf}
\endminipage\hfill
\minipage{0.33\linewidth}
    \includegraphics[trim={0cm 1.8cm 1cm 0.8cm},clip,width=0.9\linewidth]{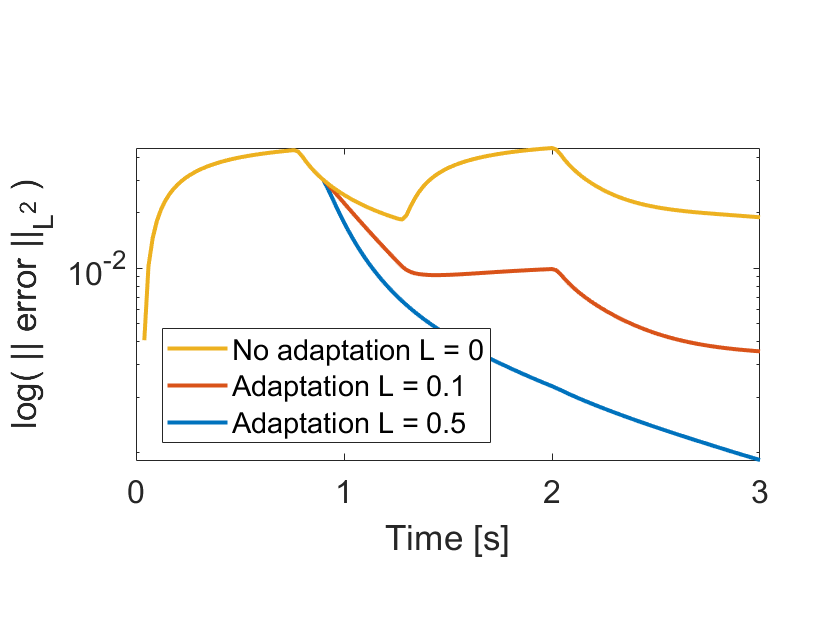}
    \caption{Temperature estimation $L^2$ error for different adaptation gains.}
    \label{fig:error l2}
\endminipage\hfill
\end{figure*}


To demonstrate the performance of the proposed adaptive observer of \eqref{eq:adaptive observer}, \eqref{eq:parameter adaptation}, we have run a simulation of an electrosurgical process. The simulation is performed on a homogeneous domain with size $4 \ \text{cm} \times 2 \ \text{cm} \times 2 \ \text{cm}$ with material properties of porcine loin tissue from \cite{El-Kebir2021d}, where mass density $\rho = 700 \ \text{kg}/\text{m}^3$, thermal conductivity $k = 0.5934 \ \text{W}/\text{kg}\cdot \text{K}$, and isochoric specific heat $c_p = 4 \ \text{kJ}/\text{kg}\cdot \text{K}$. The heat conduction is governed by the heat equation in \eqref{eq:nominal system}, and the entire domain starts at constant temperature $300 \ \text{K}$, with homogeneous Neumann boundary condition on the top surface and Dirichlet boundary condition on the other faces. Two cuts are made along the longitudinal axis on the top surface at the speed of $1 \ \text{cm}/\text{s}$. Cut 1 starts from $t=0 \ \text{s}$ at location $(1,0,0)$ and ends at point $(1.75,0,0)$, and cut 2 starts from $t=1.25 \ \text{s}$ at location $(2.25,0,0)$ and ends at point $(3,0,0)$, at $30 \ \text{W}$ output.


The heat source $q$ is a normalized 3D Gaussian distribution defined as $q(\eta) := q_0 \exp \left( -\sum_{i=1}^3 \eta_i^2 / (2 \sigma^2) \right)$,
where $\sigma = 1 \ \text{mm}$ is the radius of the electrosurgerical probe needle and $q_0$ is a normalization constant that ensures  $\int_\Omega q(\eta) \ \text{d}\eta = 1$. The domain is discretized spatially using second-order finite differencing with $\Delta x = 0.05 \ \text{cm}$, and temporally using trapezoid integration (Crank--Nicolson) with $\Delta t = 0.02 \ \text{s}$ and a Gauss--Seidel solver.

\begin{figure}[t]
    \begin{subfigure}[c]{0.49\linewidth}
    \centering
    \includegraphics[trim={0.2cm 0.28cm 1cm 0.39cm},clip,width=\linewidth]{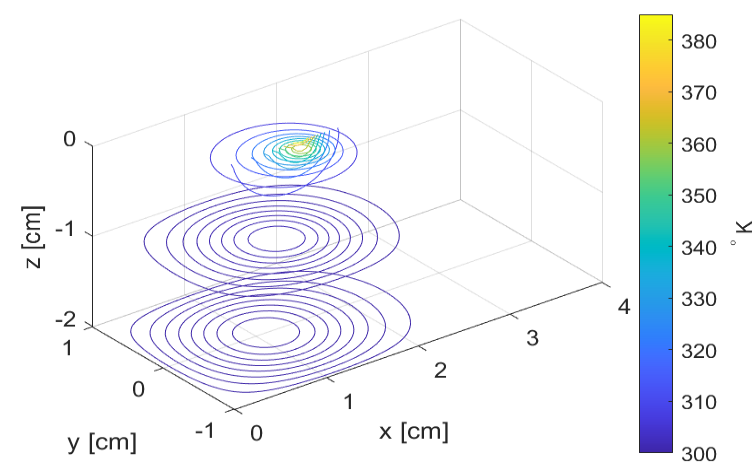}
    \caption{$t=0.54$ s.}
    \label{fig:temp result_1}
    \end{subfigure}
    \begin{subfigure}[c]{0.49\linewidth}
    \centering
    \href{https://uofi.box.com/s/ghgsv7mmjuf6ogiief1wgslhxroo2lq3}{\includegraphics[trim={0.2cm 0.2cm 1cm 0.2cm},clip,width=\linewidth]{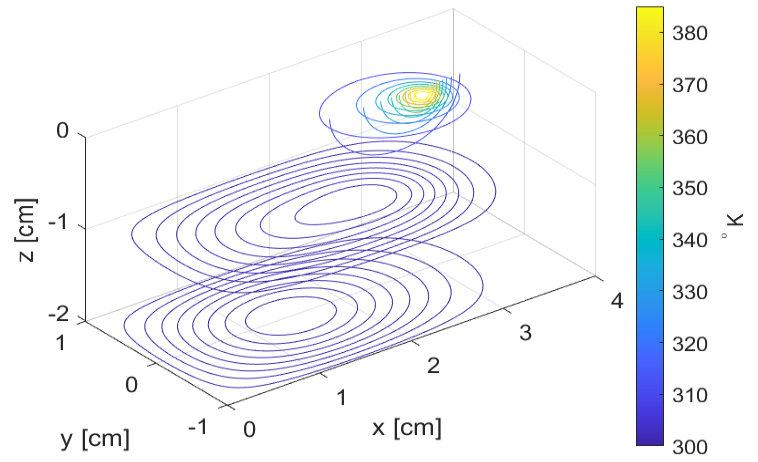}}
    \caption{$t=1.94$ s.}
    \label{fig:temp result_2}
    \end{subfigure}
    \caption{3D temperature field in Kelvin at $t= 0.54 \ \text{s}$ and $ t = 1.94 \ \text{s}$ using adaptive observer \eqref{eq:adaptive observer} with contour slices at planes $y=0$ cm, and $z = 0,-1,-2$ cm.}
\end{figure}

First, we run the simulation with the material properties listed above, with the nominal thermal diffusivity being $a=k/(\rho c_p)$. To introduce an initial parameter mismatch, we take $a_0 = 2 a$ for the observer system; by Conjecture~\ref{conj:adaptive observer convergence}, we expect the parameter mismatch to be corrected for. As mentioned in Sec.~\ref{sec:Adaptive Observer Design}, the ANRA-based thermal diffusivity is only applicable when the heat source is off, we therefore obtain $\hat{a}(t)$ for $t \in [0.75, 1.25]$ s, i.e., the period when the pen is stationary between both cuts. In addition, the heat source disturbance $w(t)$ (continuous in time, and $\gamma$-H\"older in space) is injected into the nominal system of \eqref{eq:nominal system} so as to capture thermographer noise and unmodeled dynamics. The observer gain $G(t)$ is taken to be a constant identity matrix scaled by a factor of $50$. A video\footnote{\href{https://uofi.box.com/s/ghgsv7mmjuf6ogiief1wgslhxroo2lq3}{https://uofi.box.com/s/ghgsv7mmjuf6ogiief1wgslhxroo2lq3}} of the simulated temperature field demonstrates that the adaptation law correctly detects a change in the thermal diffusivity, with the observer consequently correcting the estimated temperature field. In addition to this video, Figs.~\ref{fig:temp result_1}, \ref{fig:temp result_2} show evidence of tracking of the decreasing diffusivity, as indicated by the increased heat spread at $t = 1.94$ s.

Fig.~\ref{fig:parameter error} shows the error of the adaptive parameter during the simulation for different values of $L$. We can see that the parameter estimate converges to the correct value faster as gain $L$ in the adaptive observer increases. In the case of $L = 0.5$, the error goes to zero within one second. However, in the case of $L=0.1$, there is not enough time for the error to reach zero before the source is turned back on and parameter adaptation is suspended; given more time between cuts, this configuration would also have converged.

Referring to the nominal system run, we compute two error metrics ($\infty$-norm and $L^2$-norm of the error) over time to demonstrate convergence of the temperature estimate. In Fig.~\ref{fig:error inf} we can see the effect of the adaptive observer on the maximum temperature error. We are able to significantly reduce the error in the second cut even with a small $L$ of 0.1, and the error goes to zero with $L = 0.5$. From Fig.~\ref{fig:error l2}, we can see that the convergence rate is much higher with a larger $L$; towards the end of the simulation, most of the $L^2$ error is due to discretization errors as evidenced by Fig.~\ref{fig:error inf}. In practice, it may not be advantageous to choose exceedingly large values of $L$, since this may amplify estimation errors due to sensor noise.

%
%

\section{Conclusion}\label{sec:Conclusion}

In this work, we have considered the observation of a 3D temperature field arising during an electrosurgical procedure with a moving volumetric heat source, based only on surface temperature feedback. We have presented a nonlinear partial differential equation that governs the process, from which we derived a pointwise noncollocated observer law based on surface temperature. In this work, we have allowed for a smooth and power-bounded heat source disturbance to be present across the entire domain, including an adequate choice of feedback gain. A closed-form lower bound for the feedback gain has been derived based solely on rudimentary assumptions of the process noise, as well as the surface temperature feedback. Using this observer law, asymptotic convergence was shown.

We have also introduced a novel direct parameter estimation technique, which we termed \emph{attention-based noise-robust averaging} (ANRA) to enable real-time parameter adaptation and shown its use. We have demonstrated convergence of this adaptive observation law through numerical simulation.

\bibliographystyle{IEEEtran}
\bibliography{cdc22.bib}

%

%
%
%





\end{document}